\documentclass[a4paper, 10pt, conference]{ieeeconf}      

\usepackage[utf8]{inputenc} 
\usepackage[T1]{fontenc}    
\usepackage{url}            
\usepackage{booktabs}       
\usepackage{amsfonts}       
\usepackage{nicefrac}       
\usepackage{microtype}      
\usepackage{xcolor}         

\usepackage{graphicx}
\usepackage{epstopdf}
\usepackage{amsmath}
\usepackage{amssymb}
\usepackage{multirow}
\usepackage{tabularx}
\usepackage{colortbl}  
\usepackage{array}
\usepackage{siunitx} 
\usepackage{tcolorbox}  
\usepackage{listings}
\usepackage{mathrsfs}
\usepackage{wrapfig}
\usepackage{upgreek}
\usepackage{makecell}

\usepackage{mwe} 
\usepackage{calc} 
\usepackage{pifont} 

\definecolor{darkblue}{rgb}{0, 0.12, 0.55}
\definecolor{darkgreen}{rgb}{0, 0.55, 0.12}
\definecolor{darkred}{rgb}{0.6,0,0}
\definecolor{darkgreen}{rgb}{0,0.6,0}
\definecolor{Gray}{gray}{0.9}

\usepackage{cleveref}

\definecolor{Gray}{gray}{0.85}
\newcommand{\method}{\mbox{{ManiCM}}\xspace}

\definecolor{best}{rgb}{0.96, 0.57, 0.58}
\definecolor{second}{rgb}{0.98, 0.78, 0.57}
\definecolor{third}{rgb}{1.0, 1.0, 0.56}

\newcommand{\sdd}[2]{$#1\scriptstyle{\pm#2}$}
\newcommand{\sddbf}[2]{$\mathbf{#1\scriptstyle{\pm#2}}$}

\newcommand{\dd}[1]{$#1$}
\newcommand{\ddbf}[1]{$\mathbf{#1}$}

\usepackage{caption}
\usepackage{float}
\usepackage{subcaption}
\usepackage{placeins}
\usepackage[T1]{fontenc}

\let\titleold\title
\renewcommand{\title}[1]{\titleold{#1}\newcommand{\thetitle}{#1}}

\title{\LARGE \bf
ManiCM: Real-time 3D Diffusion Policy via Consistency Model for Robotic Manipulation
}

\author{%
  Zifeng Gao$^{1 *}$\thanks{Equal contribution. $\dagger$~ Corresponding author.},
  Guanxing Lu$^{1 *}$,
  Tianxing Chen$^2$,
  Wenxun Dai$^1$,
  Ziwei Wang$^{3}$ \\
  Chao Shang$^4$,
  Wenbo Ding$^1$,
  Yansong Tang$^{1 \dag}$ \\
  $^1$Tsinghua Shenzhen International Graduate School, Tsinghua University \\
  $^2$The University of Hong Kong,
  $^3$Nanyang Technological University \\
  $^4$Department of Automation, Tsinghua University \\
  \texttt{{lgx23@mails.,ding.wenbo@sz.,tang.yansong@sz.}tsinghua.edu.cn} \\
  \texttt{c-shang@tsinghua.edu.cn} \\
  \texttt{{konbi.gao,chentianxing2002,wxdai2001}@gmail.com, ziwei.wang@ntu.edu.sg} \\
  \url{https://ManiCM-fast.github.io}
}

\begin{document}

\maketitle
\let\thefootnote\relax\footnotetext{$*$ Equal contribution. $\dagger$ Corresponding author.}

\begin{abstract}

Diffusion models have been verified to be effective in generating complex distributions from natural images to motion trajectories. Recent diffusion-based methods show impressive performance in 3D robotic manipulation tasks, whereas they suffer from severe runtime inefficiency due to multiple denoising steps, especially with high-dimensional observations. To this end, we propose a real-time robotic manipulation model named ManiCM that imposes the consistency constraint on the diffusion process, so that the model can generate robot actions in only one-step inference.
Specifically, we formulate a consistent diffusion process in the robot action space conditioned on the point cloud input, where the original action is required to be directly denoised from any point along the ODE trajectory.
To model this process, we design a consistency distillation technique to predict the action sample directly instead of predicting the noise within the vision community for fast convergence in the low-dimensional action manifold.
We evaluate ManiCM on 31 robotic manipulation tasks from Adroit and Metaworld, and the results demonstrate that our approach accelerates the state-of-the-art method by 10 times in the average inference speed while maintaining competitive average success rate.



\end{abstract}

\section{Introduction}
\label{sec:intro}

Designing robots for diverse manipulation tasks has been highly desired in the robotic community for a long time.
In this pursuit, previous works have made great progress in exploring different architectures, e.g., perceptive models like convolutional networks~\cite{shridhar2022cliport, james2022coarse}, transformers~\cite{brohan2022rt1, brohan2023rt2}, and generative models like diffusion models~\cite{chi2023diffusionpolicy, Ze2024DP3, reuss2023goalconditionedimitationlearningusing}.
Among these choices, diffusion-based policies receive increasing attention for their capacity to model diverse high-dimensional robotic trajectories, where robot observation is provided as the condition for generation.
For instance, diffusion models have been extensively utilized across various domains as high-level procedure planners~\cite{janner2022diffuser, ajay2023is, dong2023aligndiff}, low-level motion policies~\cite{chi2023diffusionpolicy, xian2023chaineddiffuser, ma2024hierarchical, dong2024diffuserlite}, 
teacher models~\cite{lu2024manigaussian, ze2023gnfactor}, and data synthesis engines~\cite{lu2023synther}, demonstrating impressive performance in diverse robotic tasks.

\begin{figure}
    \centering
    \vspace{-1.1em}
    \includegraphics[width=1.0\linewidth]{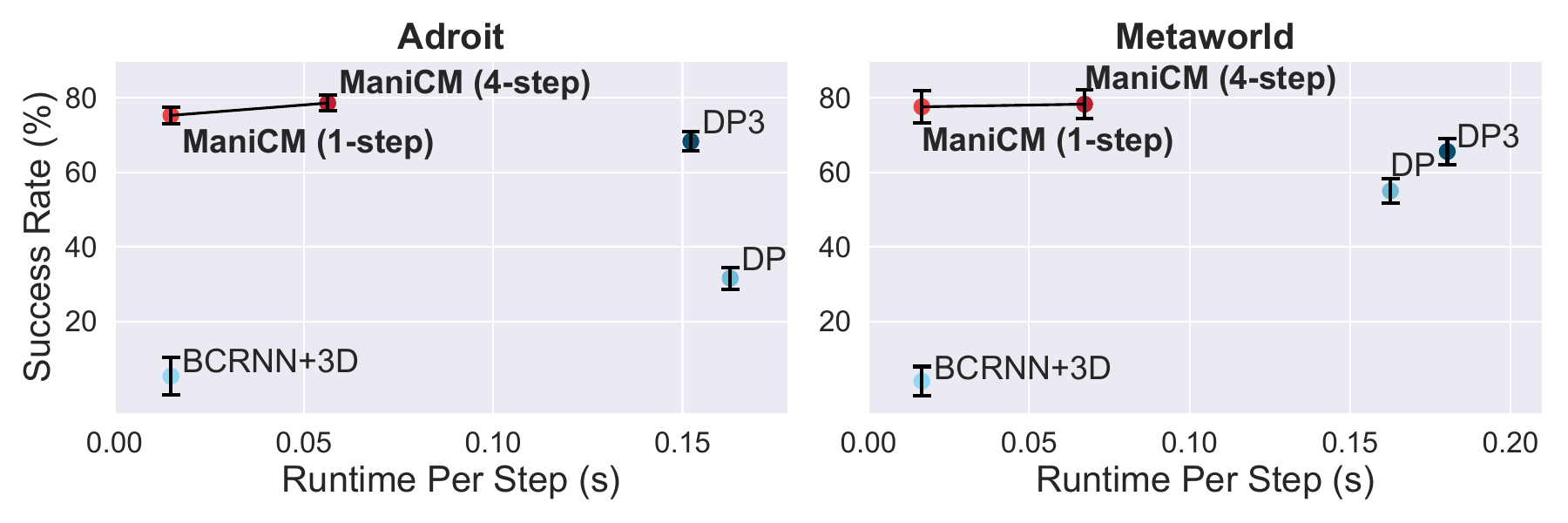}
    \vspace{-0.2in}
    \caption{\small \textbf{Trade-off Between Efficiency and Effectiveness.} We present ManiCM, a real-time 3D diffusion policy by imposing the consistency constraint on the diffusion process. BCRNN+3D~\cite{mandlekar2021matters} is a state-of-the-art perceptive model-based behavior cloning agent augmented with 3D point cloud input. DP~\cite{chi2023diffusionpolicy} and DP3~\cite{Ze2024DP3} are the state-of-the-art diffusion-based manipulation agents.
    ManiCM achieves a decision-making runtime of 16ms, which is \textbf{10}$\times$ faster than previous mainstream methods.}
    \vspace{-0.2cm}
    \label{figures/aits}
    \vspace{-0.05in}
\end{figure}

However, diffusion-based methods inevitably suffer from severe runtime inefficiency by requiring considerable sampling steps for high-quality action synthesis during inference, even with some sampling acceleration~\cite{DDIM}.
As shown in \Cref{figures/aits}, the diffusion policy (DP~\cite{chi2023diffusionpolicy}) requires $\sim$162ms per step to generate a high-quality action.
Moreover, the 3D diffusion policy (DP3~\cite{Ze2024DP3}) only achieves a decision latency of $\sim$178ms when dealing with the high-dimensional 3D point cloud input, which blocks real-time closed-loop control applications.
To address this issue, recent efforts have been proposed to speed up the inference phase of diffusion models by hierarchical sampling~\cite{xian2023chaineddiffuser, ma2024hierarchical, dong2024diffuserlite}.
Nevertheless, it is hard to determine the hierarchy across task domains that exhibit different levels of difficulty, which limits the practicality of the existing works to 3D robotic manipulation.

Recently, consistency models~\cite{cm, lcm} have been introduced to synthesize high-fidelity images with minimal steps. Inspired by this, we propose \textbf{ManiCM}, which leverages consistency distillation for real-time robotic manipulation. Different from existing methods that focus on domain-specific hierarchical planning, ManiCM achieves generalizable robot action generation in just a single inference step. Specifically, we formulate a point-cloud-conditioned diffusion process and enforce a self-consistency property along the ODE trajectory. Then, we utilize the consistency distillation technique to predict the action sample directly, based on the observation that predicting the denoised action converges faster than predicting the noise in the vision community. We evaluate ManiCM on diverse tasks from Adroit and Metaworld. The results demonstrate that our approach achieves a \textbf{10}$\times$ inference acceleration compared to state-of-the-art methods while maintaining competitive success rates (\Cref{figures/aits}). 
Our contributions can be summarized as follows:

\begin{itemize}[\itemindent = 0pt \leftmargin=0.5cm]

\item We propose a real-time 3D diffusion policy to learn robot action conditioned on the point cloud, so that the original action can be directly denoised from any point along the ODE trajectory.
                           
\item We design a manipulation consistency distillation technique to predict the action sample directly instead of predicting the noise for fast convergence to the action manifold.

\item Extensive experiments on $31$ robotic manipulation tasks from Adroit and Metaworld demonstrate that our approach accelerates the state-of-the-art method by a large margin in average inference speed while maintaining competitive average success rates.
\end{itemize}

\section{Related Work}
\label{sec:related_work}

\subsection{Diffusion Models for 3D Robotic Manipulation}
The exploration of the most effective architecture for end-to-end robotic manipulation has been a long-standing process.
The main architectures include convolutional networks~\cite{shridhar2022cliport, james2022coarse}, transformers~\cite{brohan2022rt1, brohan2023rt2}, and diffusion models~\cite{chi2023diffusionpolicy, Ze2024DP3, janner2022diffuser, ajay2023is, dong2023aligndiff, ma2024hierarchical, lu2023synther}.
Among these, diffusion models first emerged as a transformative approach in image generation~\cite{DDIM, ddpm1}, leveraging the iterative refinement of Gaussian noise into data samples.
Recently, they have been introduced to the field of embodied AI, and have made significant progress in applications that demand nuanced decision-making and adaptive control.
Specifically, previous works~\cite{chi2023diffusionpolicy, janner2022diffuser, ajay2023is} first demonstrated the potential of the diffusion model in low-dimensional control tasks.
Furthermore, the diffusion model classes~\cite{Ze2024DP3, ma2024hierarchical, 3d_diffuser_actor} are extended to more complex 3D robotic manipulation tasks, and also achieve superior performance to other traditional architectures.
However, applying diffusion models to 3D robotic manipulation faces the issue of insufficient decision-making frequency. 3D tasks involve complex spatial representations, which pose significant challenges for real-time robotic operation and control. 
Denoising processes are effective at generating high-fidelity outputs, but their iterative nature can be computationally expensive, creating a conflict between thorough exploration of the solution space and the need for fast decision-making in dynamic robotic tasks.
To solve this issue, recent methods reduce inference steps by hierarchical sampling~\cite{xian2023chaineddiffuser, ma2024hierarchical, dong2024diffuserlite}, while it is hard to determine the hierarchy across different domains. 
In this paper, we harness the power of the consistency model~\cite{cm} to improve the efficiency of diffusion models in 3D decision-making, utilizing point clouds as effective 3D representations.

\subsection{Consistency Models}

To accelerate the sampling speed of the current diffusion models (DMs)~\cite{ddpm1, ddpm2, ddpm3, sd}, the concept of consistency models (CMs)~\cite{cm} was first introduced in the image generation domain. The core idea behind the consistency models is to enforce the model to learn to map any point at any time to the origin of the probability flow ODE (PF-ODE) trajectory, \textit{i.e.}, the clean image. The consistency model facilitates efficient image generation, allowing trade-offs between extremely few inference steps and the generation quality. Recently, latent consistency models (LCMs)~\cite{lcm, lcmlora} were successfully distilled from the Stable Diffusion~\cite{sd}, achieving significant acceleration in the speed of text-conditioned image generation. Based on LCMs, \cite{ccm} introduces diverse conditional controls to manipulate the text-to-image latent consistency models. 
While initial success has been achieved in incorporating relatively simple embodied tasks with the consistency model, as explored in \cite{chen2023boosting, prasad2024consistency, ding2023consistency}, its application to complex robotic manipulation tasks with high-dimensional 3D visual conditions still remains largely unexplored.
To fill this research gap, we propose ManiCM to achieve real-time 3D diffusion policy generation.


\begin{figure}[t!]
  \centering
  \includegraphics[width=0.98\columnwidth]{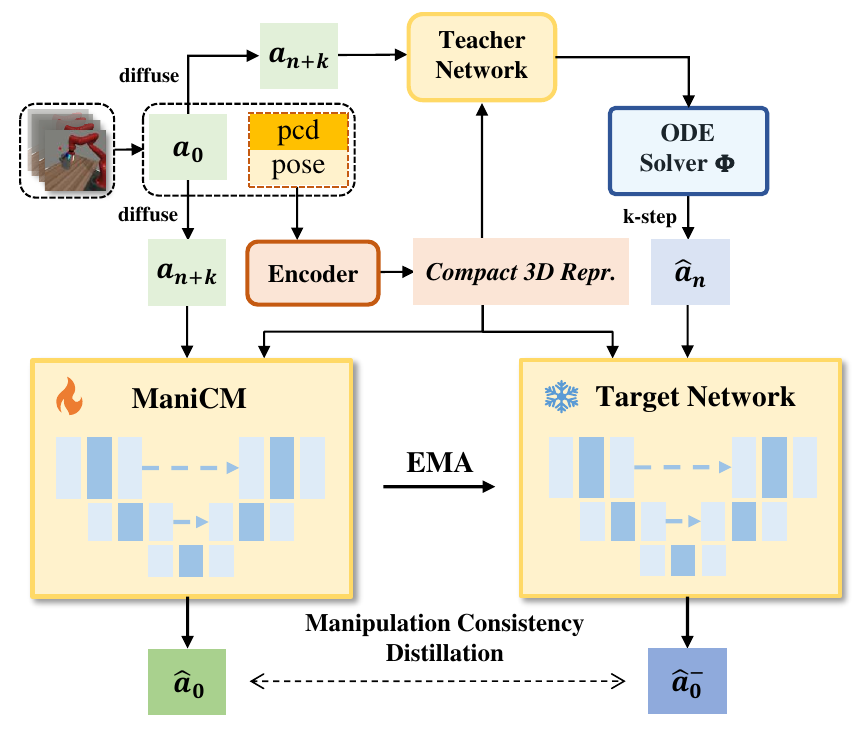} 
  \caption{\small \textbf{Overall Pipeline}. Given a raw action sequence $\boldsymbol{a}_{0}$, we first perform a forward diffusion to introduce noise over $n + k$ steps. The resulting noisy sequence $\boldsymbol{a}_{n+k}$ is then fed into both the online network and the teacher network to predict the clean action sequence. The target network uses the teacher network's $k$-step estimation results to predict the action sequence. To enforce self-consistency, a loss function is applied to ensure that the outputs of the online network and the target network are consistent.}
  \label{fig:pipeline}
  \vspace{-0.2in}
\end{figure}

\section{Approach}
\label{sec:approach}


\subsection{Problem Formulation}\label{subsec:Problem_Formulation}
The demand for robotic manipulation is of great significance in the pursuit of the general embodied agent.
To train a manipulation agent, we focus on imitation learning that learns a policy from a limited set of expert demonstrations.
The policy $\pi: \mathcal{O} \to \mathcal{A}$ maps the observation $o\in \mathcal{O}$ to the robot action $\boldsymbol{a}\in \mathcal{A}$.
The observation contains the point cloud received from an eye-to-hand RGB-D camera and the robot proprioception data.
\method uses sparse point clouds as the 3D representation. Based on the 3D visual input $o^{(t)}$ and the robot poses, the agent is required to generate the optimal action for the robot arm and grippers $\boldsymbol{a}^{(t)}$, which respectively demonstrates the change in 3D space of the end-effector followed by a normalized torque that the gripper fingers should apply. 
While existing diffusion models produce high-quality robot actions, their iterative inference limits real-time 3D manipulation. Therefore, we propose a framework based on the consistency model to enhance decision efficiency.

\subsection{ManiCM: Manipulation Consistency Model}\label{subsec:ManiCM_overall}

\noindent\textbf{3D Conditional Generation}.
To incorporate the rich spatial information embedded in the 3D point cloud, we propose to inject the point cloud representation into the diffusion policy as conditions.
Specifically, given the depth image received from a single-view RGB-D camera, we first obtain the point cloud by the extrinsic and intrinsic of the camera.
We down-sample the point cloud via farthest point sampling, in order to focus on the interest region like the robot arm and targets. Then, the sparse point cloud is encoded by a carefully designed MLP point cloud encoder, resulting in a compact 3D representation. Subsequently, we condition the original diffusion model with the compact 3D representation, which can be expressed as:
\begin{equation}
\boldsymbol{a}_{t-1}=\alpha_t\left(\boldsymbol{a}_t-\gamma_t \boldsymbol{\epsilon}_\theta\left(\boldsymbol{a}_t, t, p, q\right)\right)+\sigma_t \mathcal{N}(0, \mathbf{I}), t>0,
\label{eq:condition}
\end{equation}
where $\boldsymbol{a}_t$ denotes the noisy action at step $t$, $\boldsymbol{\epsilon}_\theta$ is the noise predicted by the diffusion model, and $p$ and $q$ denote the encoded point cloud feature and robot pose feature, respectively. The coefficients $\alpha_t$, $\gamma_t$, and $\sigma_t$ are the reverse diffusion parameters.


\begin{table*}[t]
\centering
\caption{\small \textbf{Comparisons on Success Rate and Runtime.} We evaluate $100$ episodes on $31$ challenging tasks from Adroit and Metaworld across $3$ random seeds. We report the success rates (\%) and the average runtime per step (ms) with standard deviation. The performance of our \method in one-step inference surpasses all state-of-the-art models, providing ample evidence for the effectiveness of consistency distillation. Additionally, we include two non-diffusion-based models, \textcolor{gray}{BC-GMM+3D} and \textcolor{gray}{BC-RNN+3D}, for reference.}
\label{table:comparison_with_sota} 
\renewcommand{\arraystretch}{1.1} 
\resizebox{\textwidth}{!}{ 
\begin{tabular}{l|c|ccccc|cc}
\toprule
  \multicolumn{1}{c|}{\makecell{Method}} & \makecell{NFE} & 
  \makecell{\texttt{Adroit(3)} \\ SR (\%)} & 
  \makecell{\texttt{Metaworld} \\ \texttt{Easy(18)} \\ SR (\%)} & 
  \makecell{\texttt{Metaworld} \\ \texttt{Medium(3)} \\ SR (\%)} & 
  \makecell{\texttt{Metaworld} \\ \texttt{Hard(3)} \\ SR (\%)} & 
  \makecell{\texttt{Metaworld} \\ \texttt{Very Hard(4)} \\ SR (\%)} & 
  \makecell{Avg. \\ SR (\%)} & 
  \makecell{Avg. \\ Time (ms)}\\
\midrule
    DP & \dd{10} & \dd{31.6} & \dd{77.1} & \dd{18.0} & \dd{3.0} & \ddbf{22.5} & \sdd{52.8}{3.6} & \sdd{162.6}{1.2} \\
    \textbf{Image-based ManiCM ($1$-step)} & $1$ & \ddbf{32.2} & \ddbf{77.5} & \ddbf{18.9} & \ddbf{4.3} & \dd{21.5} & \sddbf{53.1}{3.9} & \sddbf{16.0}{0.1} \\
\midrule
    Voxel-based DP & \dd{10} & \dd{35.7} & \dd{80.2} & \dd{24.5} & \ddbf{12.4} & \dd{30.8} & \sdd{57.5}{3.1} & \sdd{202.1}{1.1} \\
    \textbf{Voxel-based ManiCM ($1$-step)} & \dd{1} & \ddbf{36.4} & \ddbf{80.5} & \ddbf{24.9} & \dd{11.5} & \ddbf{31.5} & \sddbf{57.8}{3.8} & \sddbf{20.0}{0.2} \\
\midrule
    \textcolor{gray}{BC-GMM+3D} & \textcolor{gray}{1} & \textcolor{gray}{3.8} & \textcolor{gray}{4.7} & \textcolor{gray}{3.8} & \textcolor{gray}{3.4} & \textcolor{gray}{3.4} & $\textcolor{gray}{4.2}\scriptstyle\textcolor{gray}{\pm 0.4}$ & $\textcolor{gray}{15.9}\scriptstyle\textcolor{gray}{\pm 0.7}$ \\
    \textcolor{gray}{BC-RNN+3D} & \textcolor{gray}{1} & \textcolor{gray}{4.1} & \textcolor{gray}{4.5} & \textcolor{gray}{3.8} & \textcolor{gray}{3.2} & \textcolor{gray}{3.0} & $\textcolor{gray}{4.0}\scriptstyle\textcolor{gray}{\pm 0.5}$ & $\textcolor{gray}{16.6}\scriptstyle\textcolor{gray}{\pm 0.5}$ \\ 
  DP3 & \dd{10} & \dd{76.1} & \dd{91.7} & \dd{46.3} & \dd{44.8} & \dd{62.3} & \sdd{77.5}{3.8} & \sdd{177.6}{1.4} \\
  Simple DP3 & \dd{10} & \dd{73.0} & \dd{90.4} & \dd{47.3} & \dd{45.6} & \dd{56.5}  & \sdd{75.8}{3.3} & \sdd{120.6}{0.9} \\
  DP3 ($4$-step) & \dd{4} & \dd{75.3} & \dd{86.1} & \dd{39.0} & \dd{44.0} & \dd{61.3} & \sdd{72.4}{2.6} & \sdd{64.9}{0.9} \\
  Simple DP3 ($4$-step) & \dd{4} & \dd{73.0} & \dd{84.3} & \dd{37.3} & \dd{43.3} & \dd{62.0} & \sdd{70.9}{3.3} & \sdd{33.6}{0.5} \\
  \textbf{\method($1$-step)} & \dd{1} & \dd{74.9} & \dd{92.9} & \dd{47.7} & \dd{44.2} & \ddbf{65.7} & \sdd{78.5}{4.2} & \sddbf{17.3}{0.1} \\
  \textbf{\method($4$-step)} & \dd{4} & \ddbf{78.2} & \ddbf{93.0} & \ddbf{48.0} & \ddbf{46.7} & \dd{64.5} & \sddbf{79.0}{3.9} & \sdd{66.2}{0.3} \\
\bottomrule
\end{tabular}
} 
\vspace{-0.2in}
\end{table*}

\begin{figure}[ht!]
    \centering
    \includegraphics[width=\columnwidth]{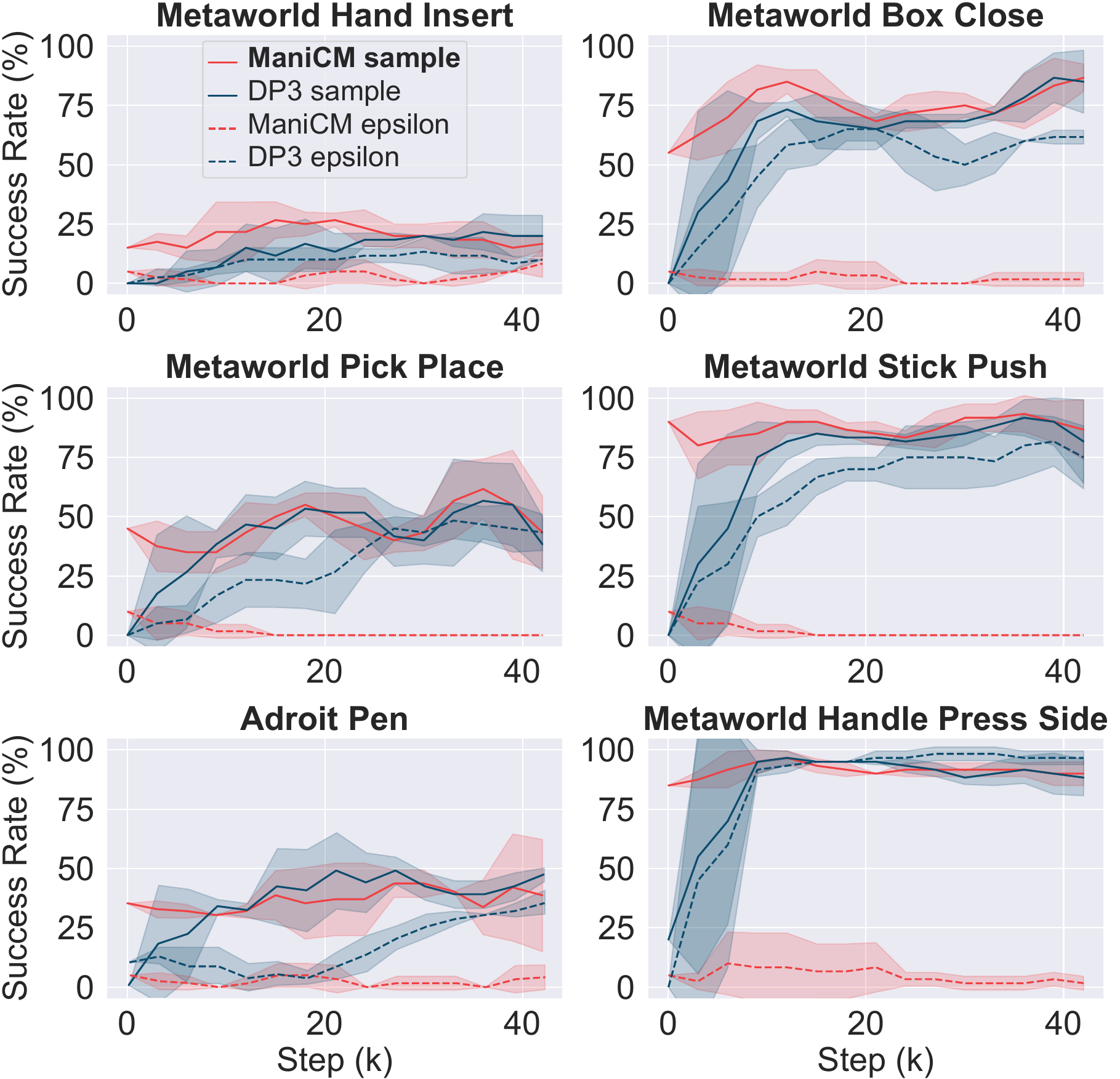}
    \caption{\small \textbf{Learning Curve.} Learning curves of \method with sample prediction vs. noise prediction. \method converges remarkably faster by predicting action sample directly than noise.}
    \label{fig:ablation_sample_vs_epsilon}
    \vspace{-0.2in}
\end{figure}

\noindent\textbf{Manipulation Self-Consistency Function}.
However, the iterative procedure in \Cref{eq:condition} raises a non-negligible computational burden, especially with high-dimensional point cloud conditions. To this end, we propose to constrain the original action to be recovered from any point along the trajectory, so that the model can generate action in only one step in the test phase:
\begin{equation}
\boldsymbol{a}_0= c_{\text{skip}}(t)\boldsymbol{a}_t + c_{\text{out}}(t)\textbf{{\textit{f}}}_{\mathbf{\theta}}\left(\boldsymbol{a}_t, t, p, q\right), t \ge 0,
\label{eq:new_condition}
\end{equation}
where $\textbf{{\textit{f}}}_{\mathbf{\theta}}$ is a manipulation consistency function that contains the accumulated $\boldsymbol{\epsilon}_\theta$ in \Cref{eq:condition} along the trajectory.
In practice, $c_{\text{skip}}(t)=\sigma_{d}^2 / (\beta^2t^2+\sigma_{d}^2)$, $c_{\text{out}}(t)=\beta t / \sqrt{(\beta^2t^2+\sigma_{d}^2)}$, where $\beta$ denotes a timestep scaling value while $\sigma_{d}$ is a balance value.
The $\textbf{{\textit{f}}}_{\mathbf{\theta}}(\cdot, \cdot)$ can be parametrized to predict the action sample or the noise epsilon. 
In mainstream diffusion generation works for image generation, the default prediction type is noise rather than sample. We speculate that while directly predicting samples may intuitively yield higher-quality generated outputs~\cite{kingma2021variational}, predicting the distribution of images is much more challenging for denoising models in high-dimensional image space.
However, the dimensionality of the robot action is quite small (approximately $28$) in manipulation tasks. In this scenario, directly learning the distribution of the samples is relatively easy and brings low variance and fast convergence, echoing \cite{Ze2024DP3}. 
More importantly, the large prediction variance caused by predicting noise will be amplified in consistency models.
Specifically, the consistency property encourages the $\textbf{{\textit{f}}}_{{\bf \theta}}(\boldsymbol{a}_{n}, t_{n})$ to match $\hat{\boldsymbol{a}}_{0} = \textbf{{\textit{f}}}_{{\bf \theta}}(\boldsymbol{a}_{0}, t_{0})$, while the original diffusion model only requires $\hat{\boldsymbol{a}}_{n}$ to be close to $\hat{\boldsymbol{a}}_{n-1}$. Thus, the variance is propagated along the ODE trajectory in consistency modeling, which ultimately blocks the convergence of noise prediction in manipulation tasks  (\Cref{fig:ablation_sample_vs_epsilon}). Therefore, we parameterize the manipulation self-consistency function $\textbf{{\textit{f}}}_{\mathbf{\theta}}(\cdot, \cdot)$ to predict the action sample rather than noise.

\subsection{Manipulation Consistency Distillation}\label{subsec:ManiCM_objective}
To train \method, we adopt the method of consistency distillation, as depicted in \Cref{fig:pipeline}. 
Here $\boldsymbol{a}_0$ represents the sequence of robotic arm actions, which comprises multiple actions, with the dimensionality determined by the specific task. $pcd$ denotes the point cloud obtained from images while $pose$ signifies the current pose of the robotic arm from the offline dataset. 
These inputs are processed through an encoder to obtain compact representations, which are then used as conditional inputs for the teacher network, online network, and target network.
Given a ground-truth expert action $\boldsymbol{a}_0$, 
we first obtain the noisy action $\boldsymbol{a}_{n+k}$ by conducting a forward diffusion operation with $n+k$ steps to add noise to $\boldsymbol{a}_0$, where $k$ is the skipping interval mentioned in \Cref{subsec:Problem_Formulation}.
Subsequently, the noisy action $\boldsymbol{a}_{n + k}$ is fed forward to a frozen teacher network and the online network to decode the clean $\hat{\boldsymbol{a}}_{0}^{*}$ and $\hat{\boldsymbol{a}}_{0}$, respectively.
In order to make sure the self-consistency property is respected, we design a target network cloned from the online network, and we expect the denoised outputs of the online network and the target network to be aligned. 
Specifically, the target network generates $\hat{\boldsymbol{a}}_{0}^{-}$ utilizing $\hat{\boldsymbol{a}}_n$. Here, $\hat{\boldsymbol{a}}_n$ is obtained by processing $\boldsymbol{a}_{n+k}$ with a $k$-step ODE solver $\boldsymbol{\Phi}$ (\textit{e.g.}, DDIM~\cite{DDIM}), which is guided by the teacher's prediction $\hat{\boldsymbol{a}}_{0}^{*}$.
In this way, we expect the consistency function $\boldsymbol{f}_{\bf{\theta}}(\cdot, \cdot)$ in~\Cref{eq:new_condition} to align the outputs $\hat{\boldsymbol{a}}_{0}$ and $\hat{\boldsymbol{a}}_{0}^{-}$.
Therefore, the consistency distillation loss is expressed as, 
\begin{equation}\label{eq:manipulation consistency loss}
    {\cal L}_{\text{MCD}}({\bf \theta}, {\bf \theta}^{-}) = \mathbb{E} \left[\left\|\textbf{{\textit{f}}}_{{\bf \theta}}(\boldsymbol{a}_{n+k}, t_{n+k}) -  \textbf{{\textit{f}}}_{{\bf \theta}^{-}}(\hat{\boldsymbol{a}}_{n}, t_n)\right\|^2_2 \right],
\end{equation}
where $\|\cdot\|_2$ is the $\ell_2$ norm. Besides, ${\bf \theta}^{-}$ is updated with the exponential moving average (EMA) of the parameter ${\bf \theta}$ defined as $
    {\bf \theta}^{-} \leftarrow \texttt{sg}(\mu{\bf \theta}^{-} + (1 - \mu) {\bf \theta})$,
where $\texttt{sg}(\cdot)$ denotes the stopgrad operation and $\mu$ satisfies $0 \leq \mu < 1$. In~\Cref{eq:manipulation consistency loss}, $\hat{\boldsymbol{a}}_{n}$ is the $k$-step estimation from $\boldsymbol{a}_{n+k}$. Therefore, the $\hat{\boldsymbol{a}}_{n}$ can be formulated as,
\begin{equation}
    \hat{\boldsymbol{a}}_{n} \leftarrow \boldsymbol{a}_{n+k} + (t_n - t_{n+k}){\mathbf{\Phi}}(\boldsymbol{a}_{n+k}, t_{n+k}),
\end{equation}
where ${\bf \Phi}(\cdot, \cdot)$ is the update function of a $k$-step ODE solver applied to PF-ODE.
As outlined in \Cref{subsec:Problem_Formulation}, we utilize EMA to update the parameters of the target network from the parameters of the online network. We refer to the pre-trained manipulation 3D diffusion model, such as DP3~\cite{Ze2024DP3}, as the "teacher network". 

Compared to other manipulation diffusion models, our ManiCM model achieves the fastest runtime (\textbf{$\sim$16ms per action}) during the inference phase by sampling high-quality actions in just one step, which is illustrated in \Cref{table:comparison_with_sota}.

\section{Experiments}
\label{sec:experiments}
\subsection{Experimental Setup}
\label{subsec:ex_setup}
\noindent\textbf{Baselines}.
We evaluate \method on 31 diverse single-task scenarios from MetaWorld~\cite{yu2021metaworld} and Adroit~\cite{adroit}, using 10 expert demonstrations per task to ensure a fair comparison with DP3~\cite{Ze2024DP3}. We also conduct multi-task experiments on 8 RLBench~\cite{james2020rlbench} tasks using 20 demonstrations and a single front camera view following ~\cite{3d_diffuser_actor}. We compare \method against state-of-the-art baselines including DP3, `Simple DP3' (a lightweight variant), Diffusion Policy (DP~\cite{chi2023diffusionpolicy}), and 3D Diffuser Actor (3dda~\cite{3d_diffuser_actor}). Following~\cite{Ze2024DP3}, our primary performance metric is the average of the top-5 success rates across 3 random seeds. To evaluate generation efficiency, we report the Number of Function Evaluations (NFE) and the average runtime per step.


\begin{table*}[t]
\small
\setlength{\tabcolsep}{12pt}
\renewcommand{\arraystretch}{1.2}
\centering
\caption{\small \textbf{Ablation on Different Diffusion Steps of the Teacher Model.} 
`SR' refers to the success rate. `$\rightarrow$' means skipping from the original training denoising steps to $k$ steps in inference with DDIM-Solver.
}
\label{table:ablation_step} 
\begin{tabular}{l|cccc|cc}
\toprule
\multirow{2}{*}{Method / DDIM Step $k$} & \multicolumn{2}{c}{\textbf{Adroit}} & \multicolumn{2}{c|}{\textbf{Metaworld}} & \multicolumn{2}{c}{\textbf{Average}} \\ 
\cmidrule{2-3}\cmidrule{4-5}\cmidrule{6-7} & Runtime & SR & Runtime & SR & Runtime & SR \\ 
\midrule
DP3 ($100\rightarrow10$) & $159.6$ & $76.1$ & $179.6$ & $77.6$ & $177.6$ & $77.5$ \\ 
DP3 ($1000\rightarrow50$) & $920.8$ & $75.5$ & $998.1$ & $78.5$ & $994.4$ & $78.2$ \\ 
\midrule
\textbf{\method} ($100\rightarrow10$) & \ddbf{16.9} & $74.9$ & $17.3$ & \ddbf{78.9} & \ddbf{17.3} & \ddbf{78.5} \\ 
\textbf{\method} ($1000\rightarrow50$) & $19.4$ & \ddbf{77.6} & \ddbf{16.9} & $75.0$ & $17.4$ & $75.3$ \\ 
\bottomrule
\end{tabular}
\vspace{-0.2in}
\end{table*}

\noindent\textbf{Implementation Details}\label{exp:implementation detail}. 
For \method, an AdamW~\cite{AdamW} optimizer with a batch size of 128 and a learning rate of 5e-5 is employed for 3K epochs training.
We also adopt a cosine decay learning rate scheduler and 500 iterations of linear warm-up.
The EMA rate is set to $\mu = 0.95$ by default and we use DDIM-Solver~\cite{DDIM} with a skipping step of $k = 10$ as ODE Solver. 
Unless otherwise specified, our experiments are conducted using $100$ training time steps. 
The teacher model employed in our experiments uses DDIM with an inference step of $10$.
For the input, we utilize an observation steps value of $2$, indicating that we use the point clouds from the two most recent time steps as conditions. Similarly, with a horizon of $4$, we predict four consecutive actions during each iteration for long-horizon tasks, while only $2$ actions are executed per prediction. To focus on the interest area, we obtain depth images with size 84$\times$84 from a single camera. The point cloud is downsampled to $512$ points using the Farthest Point Sampling (FPS) algorithm, while the dimension of agent poses and actions are determined by the specific task. Both point clouds and agent poses are encoded using an MLP encoder into embeddings of length $64$. 

All compared models are trained and evaluated on one NVIDIA RTX 4090 GPU to benchmark the decision runtime.

\begin{figure}[t!]
  \centering
  \includegraphics[width=1\linewidth]{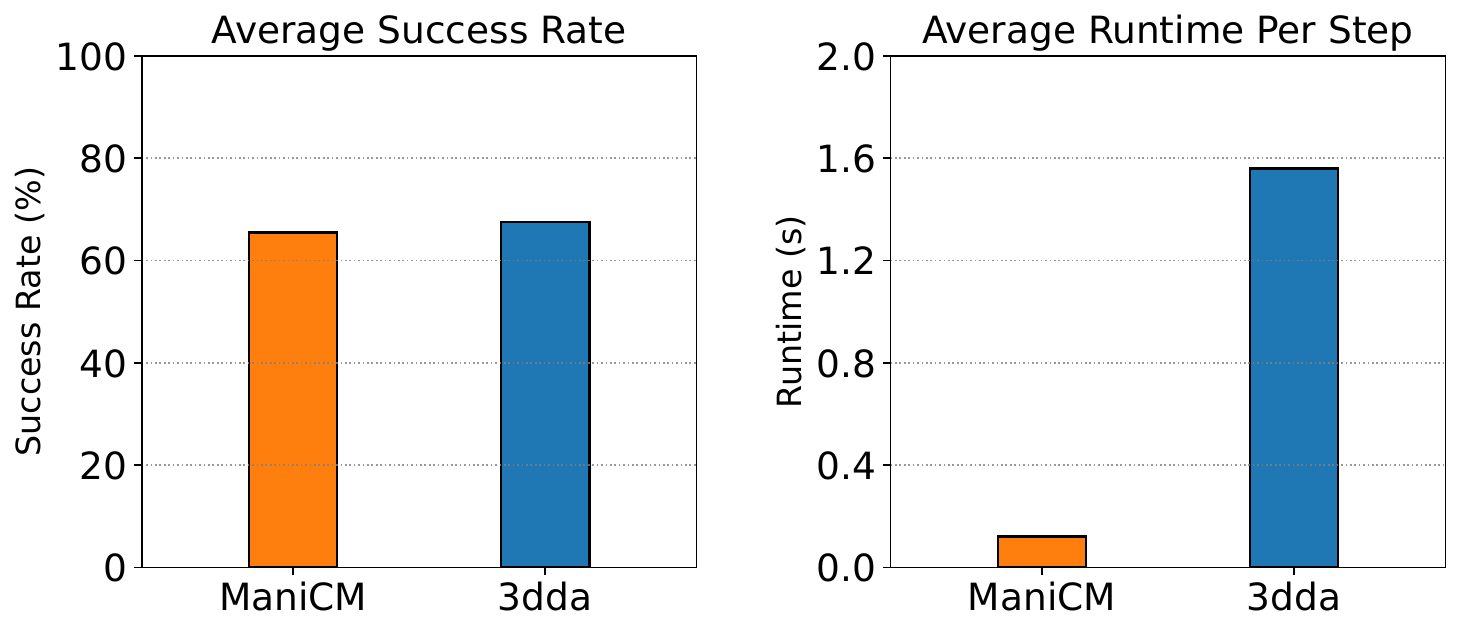}
  \caption{ 
    \textbf{Multi-Task Performance Comparison.} ManiCM balances accuracy with 3D Diffuser Actor~\cite{3d_diffuser_actor} while achieving 13.3$\times$ faster inference speeds on RLBench.
  }
  \label{fig:comparison_with_3dda}
  \vspace{-0.2in}
\end{figure}

\subsection{Comparison with the State-of-the-Art Methods}
\label{subsec:sota}
In this section, we compare \method with existing state-of-the-art methods on Adroit and Metaworld to demonstrate the effectiveness of our method.

\noindent \textbf{Comparisons on Single-Task Setup}.
\Cref{table:comparison_with_sota} illustrates the comprehensive comparisons of runtime and success rates of each model. Our \method achieves an impressive average runtime of around \textbf{16ms per step}, which is $\sim$\textbf{10}$\times$ faster than the baseline DP3. This acceleration stems from the manipulation consistency model generating robot actions in only one-step inference. Besides, the $4$-step inference version of \method also improves the efficiency by $\sim$\textbf{3}$\times$. Moreover, \method significantly surpasses the lightweight Simple DP3, indicating that reducing sampling steps is a more efficient speedup strategy than shrinking the model itself. Crucially, this massive acceleration comes without any performance drop. With only one-step inference, our \method surpasses the state-of-the-art DP3 and Simple DP3 by $1.0$\% on average, as distilling the diffusion process with consistency regularization avoids overfitting to the training demonstrations. Especially in the \texttt{Metaworld Very Hard} tasks, our method outperforms the state-of-the-art method with an absolute improvement of $3.4$\%. Furthermore, increasing the sampling steps in \method yields even better average performance ($79.0$\% vs. $78.5$\%). These results demonstrate the superiority of our \method and verify our intuition that incorporating the consistency model effectively boosts both inference speed and success rates in various robotic manipulation tasks.

\noindent \textbf{Comparisons on Multi-Task Setup}. \Cref{fig:comparison_with_3dda} compares \method with the 3D Diffuser Actor~\cite{3d_diffuser_actor} baseline on the multi-task RLBench simulation environment. The results show that \method achieves competitive performance while demonstrating a remarkable 13.3$\times$ acceleration in inference speed, highlighting our consistency model's effectiveness on a multi-task setup.

\begin{figure}[t!]
\centering
\subfloat[Slide Block\label{subfig:slide_block}]{\includegraphics[width=0.32\linewidth]{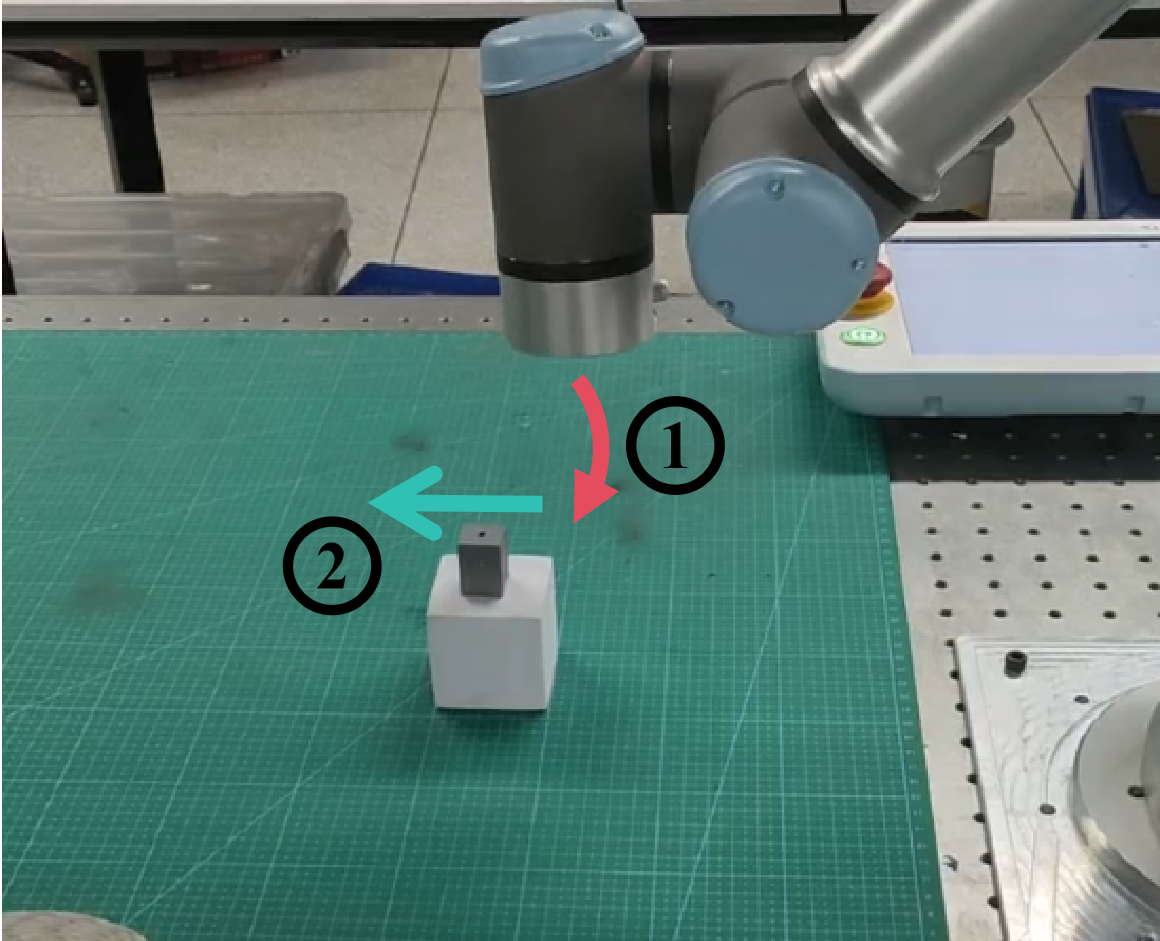}}\hfill
\subfloat[Stack Cube\label{subfig:stack_cube}]{\includegraphics[width=0.32\linewidth]{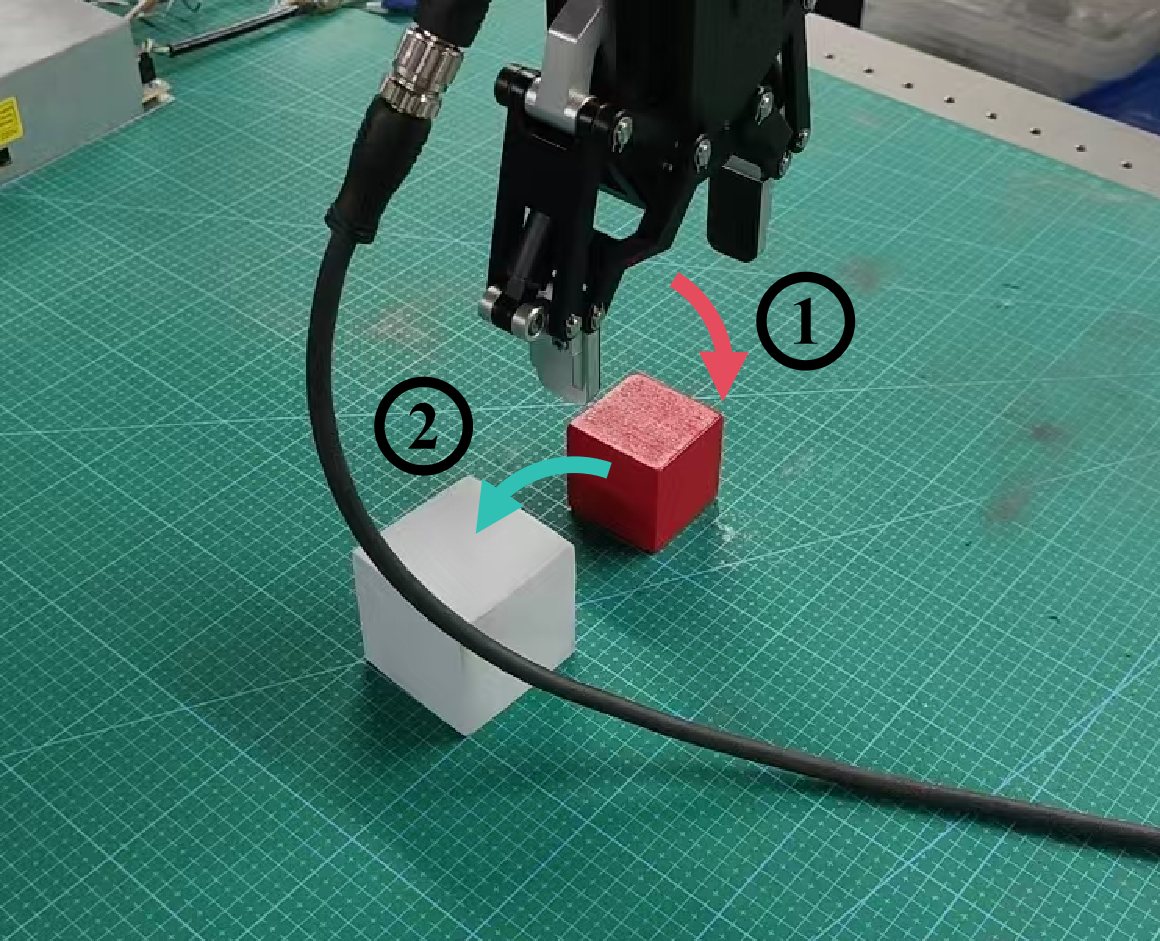}}\hfill
\subfloat[Transfer Piece\label{subfig:transfer_piece}]
{\includegraphics[width=0.32\linewidth]{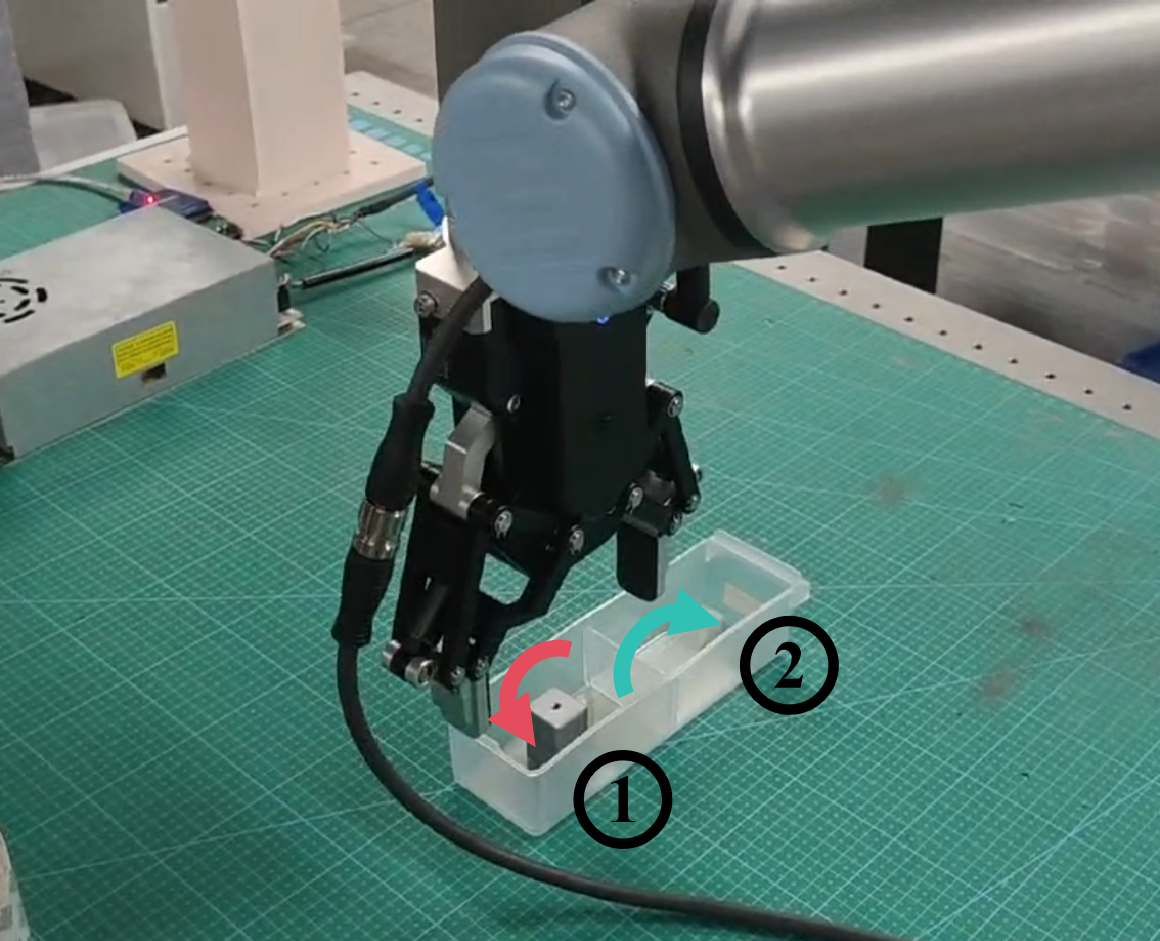}}

\caption{\textbf{Our real-world experiment setup.} We use an end effector and a DH Robotics AG-95 gripper based on UR3e arms and include three diverse manipulation tasks. A Realsense D435i camera is applied to capture visual observations.}
\label{fig:realworld_tasks}
\vspace{-0.15in}
\end{figure}

\begin{table}[t!] 
\centering
\caption{\small \textbf{Real World Experiment Results.} Each
task is evaluated with 10 trials.}
\label{table:realworld_performance}
\sisetup{
  table-number-alignment=center,
  table-format=2.0
}
\footnotesize
\setlength{\tabcolsep}{8pt}
\begin{tabular}{@{}l *{3}{S[table-format=2.0] S[table-format=3.1]}@{}}
\toprule
\textbf{Method} & 
\multicolumn{2}{c}{\textbf{\makecell{Slide Block}}} & 
\multicolumn{2}{c}{\textbf{\makecell{Stack Cube}}} & 
\multicolumn{2}{c}{\textbf{\makecell{Transfer Piece}}} \\
\cmidrule(lr){2-3} \cmidrule(lr){4-5} \cmidrule(lr){6-7}
& \textbf{SR} & \textbf{Runtime} & \textbf{SR} & \textbf{Runtime} & \textbf{SR} & \textbf{Runtime} \\
\midrule
DP3    & \ddbf{90} & 177.2 & \ddbf{80} & 181.3 & 70 & 172.4 \\
\textbf{\method} & 80 & \ddbf{16.9} & \ddbf{80} & \ddbf{17.2} & \ddbf{80} & \ddbf{16.8} \\
\bottomrule
\end{tabular}
\vspace{-0.15in}
\end{table}

\subsection{Real World Experiments}
\label{subsec:realworld}
As shown in \Cref{fig:realworld_tasks}, we evaluate ManiCM in the real world on three tasks from easy to hard with different embodiments: Slide Block, Stack Cube, and Transfer Piece.

\begin{enumerate}
  \item \textit{Slide Block}: The robot has to push the gray block off the white cube with the end-effector (see \Cref{subfig:slide_block}).

  \item \textit{Stack Cube}: The robot has to pick up the red cube and then stack it on top of the white cube with a parallel jaw gripper (see \Cref{subfig:stack_cube}).

  \item \textit{Transfer Piece}: The robot has to partially close the gripper, insert it into the left side of the box to pick the piece, and then place the piece on the right side of the box with a parallel jaw gripper. This task is more complex and requires higher precision (see \Cref{subfig:transfer_piece}).
\end{enumerate}

We use an end effector and a DH Robotics AG-95 gripper based on UR3e arms and a Realsense D435i camera to capture visual observations, with an NVIDIA RTX 4090 GPU for inference. We retain the same network architecture and input-output configuration as in the simulation environment. For each task, we conduct 10 trials to evaluate the average success rate and inference time. As demonstrated in \Cref{table:realworld_performance}, ManiCM achieves 10$\times$ faster inference speeds than DP3 while maintaining comparable success rates, validating its effectiveness in real-world setup.

\subsection{Ablation Studies}
\label{subsec:ablation}
In this section, we explore the impact of different design choices in our \method.

\noindent \textbf{Ablation on Sample Prediction and
Epsilon Prediction}.
We conduct a comparison between directly predicting the sample and predicting the noise `epsilon', which is a widely-used technique in the vision community~\cite{ddpm2}. 
From \Cref{fig:ablation_sample_vs_epsilon}, we can observe that sample prediction in the noise sampler brings faster convergence than predicting the noise epsilon in various tasks and methods. This is also observed in~\cite{Ze2024DP3}.
Besides, the performance of the noise prediction is even worse when we implement consistency distillation.
This is because predicting the action sample is more direct than training a denoising model for the low-dimensional action manifold. More importantly, the consistency distillation process will cause compound instability, so that the performance of sample prediction is much stronger than noise prediction variant ($78.5$\% vs. $24.4$\% on average).

\noindent \textbf{Ablation on Different Diffusion Steps of the Teacher Model}.
We compare our \method with DP3 under different DDIM sampling steps in \Cref{table:ablation_step}. 
We can conclude that:
1) The task success rates do not monotonically increase when taking more diffusion sampling steps in both DP3 and \method, which indicates that most of the diffusion steps are redundant. As a result, we can distill the teacher model with multiple-step inference into an online model that generates high-quality action in only one sampling step.
2) For different diffusion sampling steps, our \method can stably distill competitive capacity in task completion, where it accelerates the runtime efficiency by $\sim$\textbf{57}$\times$ when the teacher is a DDIM with $50$ inference steps.



\section{Conclusion}
\label{sec:conclusion}

In this paper, we have proposed ManiCM, a method that imposes the consistency constraint on the diffusion process, which brings 3D diffusion-based robotic manipulation to a real-time level. We design a manipulation self-consistency function that predicts the action sample directly based on 3D point cloud conditions. Subsequently, we implement manipulation consistency distillation to ensure the action is denoised from any point along the ODE trajectory within only one-step inference, which avoids the inefficient iterative denoising procedure in the original diffusion model. Extensive experiments across various task suites verify the effectiveness and efficiency of our method. 
We leave the scalability of consistency models for distilling large-scale pre-trained diffusion policies as future work.


\section*{Acknowledgement}

The authors would like to acknowledge Zhixuan Liang and Yao Mu from the University of Hong Kong for their helpful technical discussions and suggestions.

\bibliographystyle{IEEEtran} 
\bibliography{IEEEabrv,egbib}

\end{document}